\documentclass[letterpaper, 10 pt, conference]{ieeeconf}  
\pdfoutput=1

\IEEEoverridecommandlockouts                              

\usepackage{hyperref}
\usepackage{graphicx} 
\usepackage{amsmath} 
\usepackage{amssymb}  
\usepackage[utf8]{inputenc} 
\usepackage{algorithm}
\usepackage[noend]{algpseudocode}
\usepackage{color}
\usepackage{tabulary}
\usepackage{makecell}
\usepackage{booktabs}
\usepackage{wrapfig}

\title{\LARGE \bf
Using Parameterized Black-Box Priors\\ to Scale Up Model-Based Policy Search for Robotics
}

\author{Konstantinos Chatzilygeroudis and Jean-Baptiste Mouret$^{*}$
\thanks{*Corresponding author: {\tt\footnotesize jean-baptiste.mouret@inria.fr}}
\thanks{\scriptsize All authors have the following affiliations:}
\thanks{\scriptsize - Inria, Villers-lès-Nancy, F-54600, France}%
\thanks{\scriptsize - CNRS, Loria, UMR 7503, Vandœuvre-lès-Nancy, F-54500, France}%
\thanks{\scriptsize - Université de Lorraine, Loria, UMR 7503, Vandœuvre-lès-Nancy, F-54500, France}%
\thanks{\scriptsize This work received funding from the European Research Council (ERC) under the European Union's Horizon 2020 research and innovation programme (GA no. 637972, project ``ResiBots'') and the European Commission through the project H2020 AnDy (GA no. 731540).}%
}

\begin{document}

\maketitle
\thispagestyle{empty}
\pagestyle{empty}

\begin{abstract}

The most data-efficient algorithms for reinforcement learning in robotics are model-based policy search algorithms, which alternate between learning a dynamical model of the robot and optimizing a policy to maximize the expected return given the model and its uncertainties. Among the few proposed approaches, the recently introduced Black-DROPS algorithm exploits a black-box optimization algorithm to achieve both high data-efficiency and good computation times when several cores are used; nevertheless, like all model-based policy search approaches, Black-DROPS does not scale to high dimensional state/action spaces. In this paper, we introduce a new model learning procedure in Black-DROPS that leverages parameterized black-box priors to (1) scale up to high-dimensional systems, and (2) be robust to large inaccuracies of the prior information. We demonstrate the effectiveness of our approach with the ``pendubot'' swing-up task in simulation and with a physical hexapod robot (48D state space, 18D action space) that has to walk forward as fast as possible. The results show that our new algorithm is more data-efficient than previous model-based policy search algorithms (with  and without priors) and that it can allow a physical 6-legged robot to learn new gaits in only 16 to 30 seconds of interaction time.

\end{abstract}

\section{Introduction}
Robots have to face the real world, in which trying something might take seconds, hours, or even days~\cite{atkeson_no_2015}. Unfortunately, the current state-of-the-art learning algorithms (\emph{e.g.}, deep learning~\cite{lecun2015deep}) either rely on the availability of very large data sets (\emph{e.g.}, 1.2 millions labeled images in the ImageNet database~\cite{deng2009imagenet}) or only make sense in simulated environments (\emph{e.g.}, 38 days of learning for Atari games~\cite{mnih_human-level_2015}). This scarcity of data calls for algorithms that are highly data-efficient, that is, that minimize the \emph{interaction time} between the robot and the world, even if it means a considerable computation cost.

In reinforcement learning for robotics, the most data-efficient algorithms are model-based policy search algorithms~\cite{deisenroth_survey_2013,polydoros2017survey}: after each episode, the algorithm updates a model of the dynamics of the robot, then it searches for the best policy according to the model. To improve the data-efficiency, the current algorithms take the uncertainty of the model into account in order to avoid overfitting the model~\cite{deisenroth_gaussian_2015,chatzilygeroudis2017black}. The PILCO algorithm \cite{deisenroth_gaussian_2015} implements these ideas, but (1) it imposes several constraints on the reward functions and policies (because it needs to compute gradients analytically), and (2) it is a slow algorithm that cannot benefit from multi-core computers (typically about an hour to complete 15 episodes on the cart-pole benchmark)~\cite{chatzilygeroudis2017black}.

The recently introduced Black-DROPS algorithm~\cite{chatzilygeroudis2017black} is one of the first model-based policy search algorithms for robotics that is purely black-box and can extensively take advantage of parallel computations. Black-DROPS achieves similar data-efficiency to state-of-the-art approaches like PILCO (\emph{e.g.}, less than $20$\,s of interaction time to solve the cart-pole swing-up task), while being faster on multi-core computers, easier to set up, and much less limiting (\emph{i.e.}, it can use any policy and/or reward parameterization; it can even learn the reward model).
\begin{figure}[!t]
  \centering
  \includegraphics[width=0.975\linewidth]{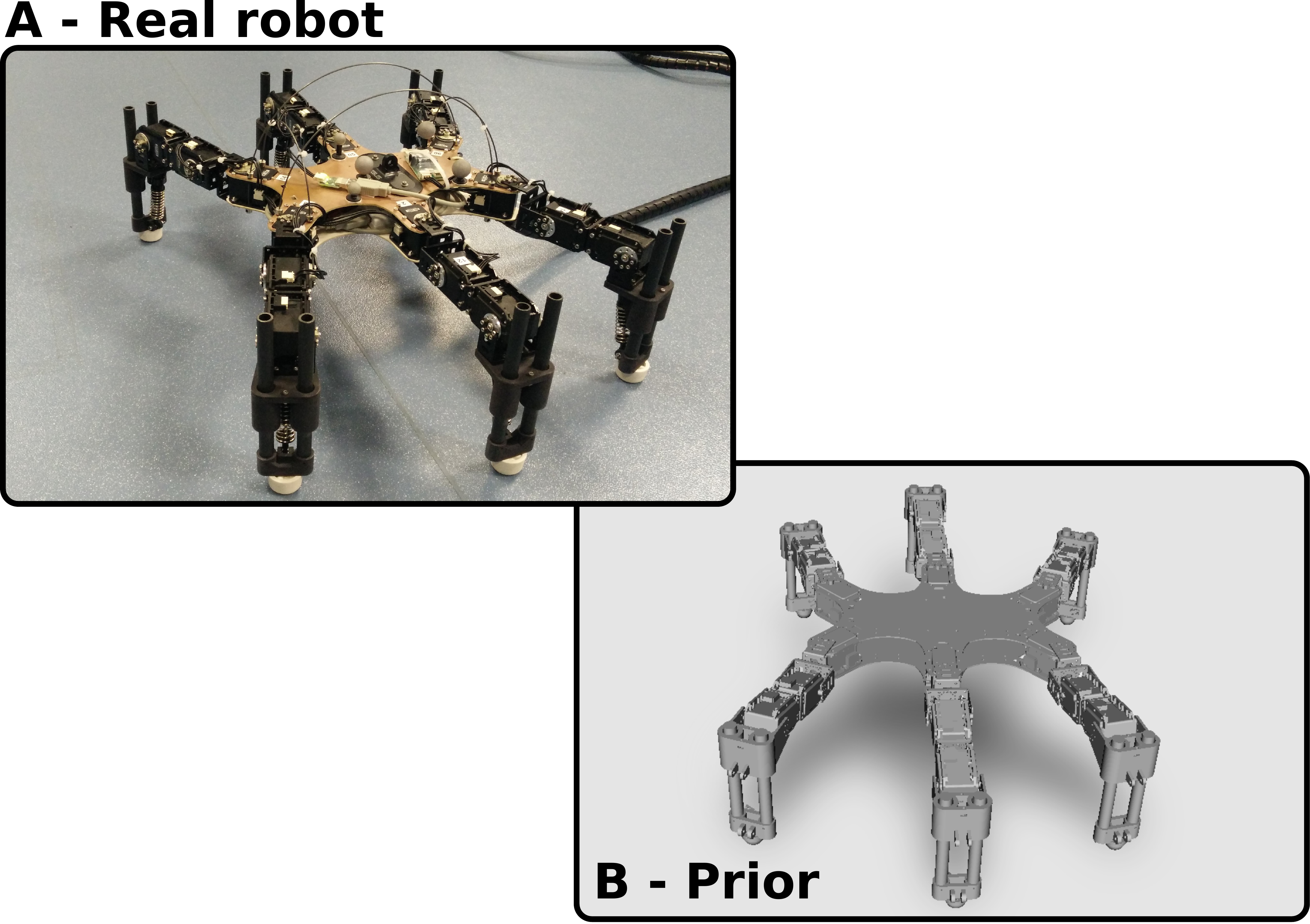}
  \vspace{-0.7em}
  \caption{\textbf{A.} The physical hexapod robot used in the experiments (48D state space and 18D action space). \textbf{B.} The simulated hexapod that is used as a prior model for our approach in the experiments.}
  \label{fig:hexapod}
  \vspace{-1.5em}
\end{figure}

However, while Black-DROPS scales well with the number of processors, the main challenge of model-based policy search is scaling up to complex problems: as the algorithm models the transition function between full state/action spaces (joint positions, environment, joint velocities, \emph{etc.}), the complexity of the model increases substantially with each new degree of freedom; unfortunately, the quantity of data required to learn a good model scales most of the time exponentially with the dimension of the state space~\cite{keogh2011curse}. As a consequence, the data-efficiency of model-based approaches greatly suffers from the increase of the dimensionality of the model. In practice, model-based policy search algorithms can currently be employed only with simple systems up to 10-15D state and action space combined (\emph{e.g.}, double cart-pole or a simple manipulator).

One way of tackling the problem raised by the ``curse of dimensionality'' is to use prior information about the system that is modeled; for instance, dynamic simulators of the robot can be effective priors and are often available.
The ideal model-based policy search algorithm with priors for robotics should, therefore:

\begin{itemize}
  \item scale to high dimensional and complex robots (\emph{e.g.}, walking or soft robots);
  \item take advantage of multi-core architectures to speed-up computation times;
  \item perform the search in the full policy space (\emph{i.e.}, the more real trials, the better expected reward);
  \item make as few assumptions as possible about the type of robot and the prior information (\emph{i.e.}, require no specific structure or differentiable models);
  \item be able to select among several prior models or to tune the prior model.
\end{itemize}

A few algorithms leverage prior information to speed-up learning on the real system~\cite{cutler_efficient_2015,lee2017gp,saverianodata,bischoff_policy_2014,cully2015robots,marco17virtualvsreal}, but none of them fulfills all of the above properties. In this paper, we propose a novel, purely black-box, flexible and data-efficient model-based policy search algorithm that combines ideas from the Black-DROPS algorithm, from simulation-based priors, and from recent model learning algorithms~\cite{nguyen2010using,camoriano2016incremental}. We show that our approach is capable of learning policies in about 30 seconds to control a damaged physical hexapod robot (48D state space, 18D action space) and outperforms state-of-the-art model-based policy search algorithms without (PILCO~\cite{deisenroth_gaussian_2015}, Black-DROPS~\cite{chatzilygeroudis2017black}) and with priors (PILCO with priors~\cite{cutler_efficient_2015}), as well as prior-based Bayesian optimization (IT\&E~\cite{cully2015robots}).


\section{Background}

\subsection{Policy Search for Robotics}

Model-free policy search (PS) methods have been successful in robotics as they can easily be applied in high-dimensional continuous state-action RL problems~\cite{deisenroth_survey_2013,lillicrap_continuous_2015,schulman2015trust}. The PoWER algorithm~\cite{kober2011power} uses probability-weighted averaging, which has the property of following the natural gradient without computing it. The PI$^2$~\cite{theodorou2010generalized} algorithm has very similar performance with PoWER, but puts no constraint on the reward function. Natural Evolution Strategies (NES)~\cite{wierstra2014natural} and Covariance Matrix Adaptation ES (CMA-ES)~\cite{hansen2001completely} families of algorithms are population-based black-box optimizers that iteratively update a search distribution by calculating an estimated gradient on the distribution parameters (mean and covariance). At each generation, they sample a set of policy parameters and rank them based on their expected return. NES performs gradient ascent along the natural gradient, whereas CMA-ES updates the distribution by exploiting the technique of evolution paths.

Although, model-free policy search methods are promising, they require a few hundreds or thousands of episodes to converge to good solutions~\cite{deisenroth_survey_2013,polydoros2017survey}. The data-efficiency of such methods can be increased by learning the model (\emph{i.e.}, transition and reward function) of the system from data and inferring the optimal policy from the model~\cite{deisenroth_survey_2013,polydoros2017survey}. For example, state-of-the-art model-free policy gradient methods (\emph{e.g.}, TRPO~\cite{schulman2015trust} or DDPG~\cite{lillicrap_continuous_2015}) require more than $500$\,$s$ of interaction time to solve the cart-pole swing-up task~\cite{lillicrap_continuous_2015} whereas state-of-the-art model-based policy search algorithms (\emph{e.g.}, PILCO or Black-DROPS) require less than $20$\,$s$~\cite{chatzilygeroudis2017black,deisenroth_gaussian_2015}. Probabilistic models have been more successful than deterministic ones, as they provide an estimate about the uncertainty of their approximation which can be incorporated into long-term planning~\cite{deisenroth_gaussian_2015,chatzilygeroudis2017black,polydoros2017survey,deisenroth_survey_2013}.

Black-DROPS~\cite{chatzilygeroudis2017black} and PILCO~\cite{deisenroth_gaussian_2015} are two of the most data-efficient model-based policy search algorithms for robot control. They essentially differ in how they use the uncertainty of the model and in how they optimize the policy given the model: PILCO uses moment matching and analytical gradients~\cite{deisenroth_gaussian_2015}, whereas Black-DROPS uses Monte-Carlo rollouts and a black-box optimizer.

Black-DROPS adds two main benefits to PILCO: (1) any reward function or policy parameterization can be used (including non-differentiable policies like finite automata), and (2) it is a highly-parallel algorithm that takes advantages of multi-core computers. 
Black-DROPS achieves similar data-efficiency to PILCO and escapes local optima faster in standard control benchmarks (inverted pendulum and cart-pole swing-up)~\cite{chatzilygeroudis2017black}. It was also able to learn from scratch a high dimensional policy (neural network with 134 parameters) in only 5-6 trials on a physical low-cost manipulator~\cite{chatzilygeroudis2017black}.

\subsection{Accelerating Policy Search using Priors}

Model-based policy search algorithms reduce the required interaction time, but for more complex or higher dimensional systems, they still require dozens or even hundreds of episodes to find a working policy; in some systems, they might also fail to find any good policy because of the inevitable model errors and biases~\cite{sutton1998reinforcement}.

One way to reduce the interaction time without learning models is to begin with a meaningful initial policy (coming from demonstration or simulation) and then search locally to improve it. Usually this is done by human demonstration and movement primitives~\cite{kober2010imitation}: a human either tele-operates or moves the robot by hand trying to achieve the task and then a model-free RL method is applied to improve the initial policy~\cite{kober2011power,stulp2013robot}. However, these approaches still suffer from the data inefficiency of model-free approaches and require dozens or hundreds of episodes to find good policies.

Another way to reduce the interaction time in model-free approaches is to pre-compute archives/libraries of policies/controllers~\cite{cully_behavioral_2013,majumdar2017funnel} and then search online for the one that works best on the real system~\cite{cully2015robots,antonova2016sample}. The Intelligent Trial-and-Error (IT\&E) algorithm~\cite{cully2015robots} first uses an evolutionary algorithm called MAP-Elites~\cite{mouret_illuminating_2015,vassiliades2017using} off-line to create an archive of diverse and locally high-performing behaviors and then utilizes a modified version of Bayesian optimization (BO)~\cite{shahriari_taking_2016} to quickly find a compensatory behavior. Although IT\&E can allow, for instance, a damaged 6-legged robot to find a new gait in about a dozen trials (less than 2 minutes) and a robotic arm to overcome several blocked joints in a few minutes, it is not searching in the full policy space and as such there is no guarantee that the optimal policy can be found.

Reducing the interaction time in model-based policy search can be achieved by using priors on the models~\cite{cutler_efficient_2015,lee2017gp,saverianodata,bischoff_policy_2014,ko2007gaussian}; \emph{i.e.}, starting with an initial guess of the dynamics and then learning the residual model. PILCO with priors~\cite{cutler_efficient_2015} and PI-REM~\cite{saverianodata} are closely related as they both use the policy search procedure of PILCO. PILCO with priors uses simulated data to create a Gaussian process prior, whereas PI-REM uses analytic equations for the prior model. The main limitation of PILCO with priors is that it implicitly requires the task to be solved in the prior model with PILCO (in order to get the speed-up shown in the original paper~\cite{cutler_efficient_2015}). GP-ILQG~\cite{lee2017gp} also learns the residual model like PI-REM and then uses a modified version of ILQG~\cite{todorov2005generalized} to find a policy given the uncertainties of the model. GP-ILQG, however, requires the prior model to be differentiable.

\subsection{Model Identification and Learning}

The traditional way of exploiting analytic equations is model identification~\cite{hollerbach2016model}. Most approaches for model identification rely on two main ingredients: (a) proper excitation of the system~\cite{gautier1992exciting,hollerbach2016model,aghili2007modular} and (b) parametric models. 
Recently, Xie et. al.~\cite{xie2016model} proposed a method that combines model identification and RL. More specifically, their approach relies on a Model Predictive Control (MPC) scheme with optimistic exploration on a parametric model that is estimated from the collected data using least-squares.

However, these approaches assume that the analytical equations can fully capture the system, which is often not the case when dealing with unforeseen effects like, for example, complex friction effects or when there exists severe model mismatch (\emph{i.e.}, no parameters can explain the data) like, for instance, when the robot is damaged.

A few methods have been proposed to combine model identification and model learning~\cite{nguyen2010using,camoriano2016incremental}. Nevertheless, these methods are based on the manipulator equation exploiting it in different ways and it is not straight-forward how they can be used with more complicated robots that involve complex collisions and contacts (\emph{e.g.}, walking or complex soft robots).


\begin{algorithm}[t]
  \caption{Model-based policy search with priors}
  \label{algo:model_ps_priors}
  \begin{algorithmic}[1]
   \State Optimize $\boldsymbol{\theta}^*$ on the prior model according to $J(\boldsymbol{\theta})$ and the initial reward function $r$
   \State Apply policy $\pi_{\boldsymbol{\theta}^*}$ on the robot and record data
   \Repeat \label{algo:model_ps_priors:init}
    \State Learn the immediate reward function $r$ from the gathered data --- if necessary
    \State Learn a model that approximates the actual underlying system's dynamics using the gathered data \textit{and the prior model}
    \State Optimize $\boldsymbol{\theta}^*$ on the model according to $J(\boldsymbol{\theta})$ and the (learned) reward function $r$
    \State Apply policy $\pi_{\boldsymbol{\theta}^*}$ on the robot and record data
   \Until{Task is solved}\label{algo:model_ps_priors:end}
  \end{algorithmic}
  \end{algorithm}

\section{Problem Formulation}

\noindent We consider dynamical systems of the form:
\begin{align}
 \mathbf{x}_{t+1} = \mathbf{x}_t + F(\mathbf{x}_t,\mathbf{u}_t) + \mathbf{w}
\end{align}
with continuous-valued states $\mathbf{x}\in\mathbb{R}^E$ and controls $\mathbf{u}\in\mathbb{R}^U$, i.i.d. Gaussian system noise $\mathbf{w}$, and unknown transition dynamics $F$. We assume that we have an initial guess of the dynamics, the function $M(\mathbf{x}_t,\mathbf{u}_t)$, that may not be accurate either because we do not have a very precise model of our system (\emph{i.e.}, what is called the \textit{``reality-gap''}~\cite{mouret201720}) or because the robot is damaged in an unforeseen way (\emph{e.g.}, a blocked joint or faulty motor/encoder)~\cite{cully2015robots,chatzilygeroudis2016resetfree}.

Contrary to previous works~\cite{lee2017gp,nguyen2010using,camoriano2016incremental}, we assume no structure or specific properties of our initial dynamics model $M$ (\emph{i.e.}, we treat it as a black-box function), other than it has some tunable parameters, $\boldsymbol{\phi}_M$, which change its behavior. Examples of these parameters can be some optimization parameters (\emph{e.g.}, type of optimizer) of a dynamic simulator involving contacts and collisions or some internal parameters of the robot (\emph{e.g.}, masses of the bodies). Finally, we add a non-parametric model, $f$ (with associated hyper-parameters $\boldsymbol{\phi}_K$), to model whatever is not possible to capture with $M$:
\begin{align}
 \mathbf{x}_{t+1} = \mathbf{x}_t + M(\mathbf{x}_t,\mathbf{u}_t, \boldsymbol{\phi}_M) + f(\mathbf{x}_t,\mathbf{u}_t, \boldsymbol{\phi}_K) + \mathbf{w}
\end{align}
Our objective is to find a deterministic \textit{policy} $\pi$, $\mathbf{u} = \pi(\mathbf{x}|\boldsymbol{\theta})$ that maximizes the \textit{expected long-term reward} when following policy $\pi$ for $T$ time steps:
\begin{align}
  \label{eq:reward_j}
  J(\boldsymbol{\theta}) = \mathbb{E} \Bigg[\sum_{t=1}^{T}r(\mathbf{x}_t) \Big| \boldsymbol{\theta} \Bigg]
\end{align}
where $r(\mathbf{x}_t)$ is the immediate reward of being in state $\mathbf{x}_t$.
We assume that $\pi$ is a function parameterized by $\boldsymbol{\theta}\in\mathbb{R}^{\Theta}$.

In model-based policy search with priors, we begin by optimizing the policy on the prior model (that is, \emph{there is no prior information on the policy parameters}) and applying it on the real system to gather the initial data. Afterwards, a loop is iterated where we first learn a model using the prior model and the collected data and then optimize the policy given this newly learned model (Algo.~\ref{algo:model_ps_priors}). Finally, the policy is applied on the real system, more data is collected and the loop re-iterates until the task is solved.

\section{Approach}

\subsection{Gaussian processes with the simulator as the mean function}
We would like to have a model $\hat{F}$ that approximates as accurately as possible the unknown dynamics $F$ of our system given some initial guess, $M$. 
We rely on Gaussian processes (GPs) to do so as they have been successfully used in many model-based reinforcement learning approaches~\cite{deisenroth_gaussian_2015,chatzilygeroudis2017black,engel_reinforcement_2005,nguyen2011model,deisenroth_survey_2013,chatzilygeroudis2016resetfree,polydoros2017survey}.
A GP is an extension of the multivariate Gaussian distribution to an infinite-dimension stochastic process for which any finite combination of dimensions will be a Gaussian distribution~\cite{rasmussen2006gaussian}. 

As inputs, we use tuples made of the state vector $\mathbf{x}_t$ and the action vector $\mathbf{u}_t$, that is, $\mathbf{\tilde{x}}_t = (\mathbf{x}_t,\mathbf{u}_t)\in\mathbb{R}^{E+U}$;
as training targets, we use the difference between the current state vector and the next one: $\mathbf{\Delta}_{\mathbf{x}_t} = \mathbf{x}_{t+1}-\mathbf{x}_t\in\mathbb{R}^E$. We use $E$ independent GPs to model each dimension of the difference vector $\mathbf{\Delta}_{\mathbf{x}_t}$.
Assuming $D_{1:t} = \{F(\mathbf{\tilde{x}}_1),...,F(\mathbf{\tilde{x}}_t)\}$ is a set of observations and $M(\mathbf{\tilde{x}})$ being the simulator function (\emph{i.e.}, our initial guess of the dynamics --- tunable or not; we drop the $\boldsymbol{\phi}_M$ parameters here for brevity), we can query the GP at a new input point $\mathbf{\tilde{x}}_*$:
\begin{align}
  \label{eq:gp}
  p(\hat{F}(\mathbf{\tilde{x}}_*)|D_{1:t},\mathbf{\tilde{x}}_*) = \mathcal{N}(\mu(\mathbf{\tilde{x}}_*),\sigma^{2}(\mathbf{\tilde{x}}_*))
\end{align}
The mean and variance predictions of this GP are computed using a kernel vector $\pmb{k} = k(D_{1:t},\mathbf{\tilde{x}}_*)$, and a kernel matrix $K$, with entries $K^{ij} = k(\mathbf{\tilde{x}}_i,\mathbf{\tilde{x}}_j)$:
\begin{align}
  \label{eq:gp_detail}
  &\mu(\mathbf{\tilde{x}}_*) = M(\mathbf{\tilde{x}}_*)+\pmb{k}^{T}K^{-1}(D_{1:t}-M(\mathbf{\tilde{x}}_{1:t}))\nonumber\\
  &\sigma^{2}(\mathbf{\tilde{x}}_*) = k(\mathbf{\tilde{x}}_*,\mathbf{\tilde{x}}_*)-\pmb{k}^{T}K^{-1}\pmb{k}
\end{align}
The formulation above allows us to combine observations from the simulator and the real-world smoothly. In areas where real-world data is available, the simulator's prediction will be corrected to match the real-world ones. On the contrary, in areas far from real-world data, the predictions resort to the simulator~\cite{cully2015robots,lee2017gp,chatzilygeroudis2016resetfree}.

This model learning procedure has been used in several articles~\cite{ko2007gaussian,nguyen2010using,nguyen2011model} and in particular to learn the cumulative reward model for a BO procedure highlighted in the IT\&E approach~\cite{cully2015robots}. GP-ILQG~\cite{lee2017gp} and PI-REM~\cite{saverianodata} formulate a similar model learning procedure for optimal control (under model uncertainty) and policy search respectively. GP-ILQG additionally assumes that the prior model $M$ is differentiable, which is not always true and might be too slow to perform via finite differences (\emph{e.g.}, when using black-box simulators for $M$). PILCO with priors~\cite{cutler_efficient_2015} utilizes a similar scheme but assumes that the prior model $M$ is a GP learned from simulation data that is gathered from running PILCO on the prior system.

We use the exponential kernel with automatic relevance determination~\cite{rasmussen2006gaussian} ($\boldsymbol{\phi}_K$ are the kernel hyper-parameters).
When searching for the best kernel hyper-parameters through Maximum Likelihood Estimation (MLE) for a GP with a non-tunable mean function $M$, we seek to maximize~\cite{rasmussen2006gaussian}:
%
\begin{align}
  \label{eq:lik}
	p(&D_{1:t}|\mathbf{\tilde{x}}_{1:t}, \boldsymbol{\phi}_K) =\nonumber\\&\frac{1}{\sqrt{(2\pi)^t |K|}} e^{-\frac{1}{2} (D_{1:t}-M(\mathbf{\tilde{x}}_{1:t}))^T K^{-1} (D_{1:t}-M(\mathbf{\tilde{x}}_{1:t}))}
\end{align}
The gradients of this likelihood function can be analytically computed, which makes it possible to use any gradient based optimizer (we use Rprop~\cite{blum2013optimization}). Since we have $E$ independent GPs, we have $E$ independent optimizations. We use the limbo C++11 library for GP regression~\cite{cully2016limbo}.

\subsection{Mean functions with tunable parameters}
\label{sec:gp_mi}

We would like to use a mean function $M(\mathbf{\tilde{x}},\boldsymbol{\phi}_M)$, where each vector $\boldsymbol{\phi}_M\in\mathbb{R}^{n_M}$ corresponds to a different prior model of our system (\emph{e.g.}, different lengths of links). Searching for the $\boldsymbol{\phi}_M$ that best matches the observations can be seen as a model identification procedure, which could be solved via minimizing the mean squared error; nevertheless, the GP framework allows us to jointly optimize for the kernel hyper-parameters and the mean parameters, which allows the modeling procedure to balance between non-parametric and parametric modeling. We can easily extend Eq.~\eqref{eq:lik} to include parameterized mean functions:
%
\begin{align}
  \label{eq:lik_mean}
	p(&D_{1:t}|\mathbf{\tilde{x}}_{1:t}, \boldsymbol{\phi}_K, \boldsymbol{\phi}_M) =\nonumber\\&\frac{1}{\sqrt{(2\pi)^t |K|}} e^{-\frac{1}{2} (D_{1:t}-M(\mathbf{\tilde{x}}_{1:t}, \boldsymbol{\phi}_M))^T K^{-1} (D_{1:t}-M(\mathbf{\tilde{x}}_{1:t}, \boldsymbol{\phi}_M))}
\end{align}
This time, even though we have $E$ independent GPs (one for each output dimension), all of them need to share the same mean parameters $\boldsymbol{\phi}_M$ (contrary to the kernel parameters, which are typically different for each dimension), because the model of the robot should be consistent in all of the output dimensions. Thus, we have to jointly optimize for the mean parameters and the kernel hyper-parameters of all the GPs. Since most dynamic simulators are not differentiable (or too slow to differentiate by finite differences), we cannot resort to gradient-based optimization to optimize Eq.~\eqref{eq:lik_mean} jointly for all the GPs. A black-box optimizer like CMA-ES~\cite{hansen2001completely} could be employed instead, but this optimization was too slow to converge in our preliminary experiments.

To combine the benefits of both gradient-based and gradient-free optimization, we use gradient-based optimization for the kernel hyper-parameters (since we know the analytical gradients) and black-box optimization for the mean parameters. Conceptually, we would like to optimize for the mean parameters, $\boldsymbol{\phi}_M$, given the optimal kernel hyper-parameters for each of them. Since we do not know them before-hand, we use two nested optimization loops: (a) an outer loop where a gradient-free local optimizer searches for the best $\boldsymbol{\phi}_M$ parameters (we use a variant of the Subplex algorithm~\cite{rowan1990functional} provided by NLOpt~\cite{johnsonnlopt} for continuous spaces and exhaustive search for discrete ones), and (b) an inner optimization loop where given a mean parameter vector $\boldsymbol{\phi}_M$, a gradient-based optimizer searches for the best kernel hyper-parameters (each GP is independently optimized since $\boldsymbol{\phi}_M$ is fixed in the inner loop) and returns a score that corresponds to $\boldsymbol{\phi}_M$ for the optimal $\boldsymbol{\phi}_K$ (Algo.~\ref{algo:gp_mi}).

\begin{algorithm}[t]
\caption{GP-MI Learning process}
\label{algo:gp_mi}
\begin{algorithmic}[1]
  \Procedure{GP-MI}{$D_{1:t}$}
  \State Optimize $\boldsymbol{\phi}_M^*$ according to \textsc{EvaluateModel($\boldsymbol{\phi}_M$, $D_{1:t}$)} using a gradient-free local optimizer
  \State
  \Return $\boldsymbol{\phi}_M^*$
  \EndProcedure
  \Procedure{EvaluateModel}{$\boldsymbol{\phi}_M$, $D_{1:t}$}
  \State Initialize $E$ GPs $f_1,\dots,f_E$ as $f_i(\mathbf{\tilde{x}})\sim\mathcal{N}(M_i(\mathbf{\tilde{x}},\boldsymbol{\phi}_M), k_i(\mathbf{\tilde{x}}, \mathbf{\tilde{x}}))$\Comment{{\scriptsize$M^i$ queries $M$ and returns the $i$-th element of the return vector, $k_i$ is the kernel function of the $i$-th GP}}
  \For{$i$ from $1$ to $E$}\Comment{{\scriptsize This can also be done in parallel}}
  \State Optimize the kernel hyper-parameters, $\boldsymbol{\phi}_K^i$, of $f_i$ given $D^i_{1:t}$ assuming $\boldsymbol{\phi}_M$ fixed \Comment{{\scriptsize$D^i_{1:t}$ is the $i$-th column of $D_{1:t}$}}
  \State $\text{lik}_i = p(D^i_{1:t}|\mathbf{\tilde{x}}_{1:t}, \boldsymbol{\phi}_K^i, \boldsymbol{\phi}_M)$ \Comment{Eq.~\eqref{eq:lik_mean}}
  \EndFor
  \State
  \Return $\sum_{i=1}^E\text{lik}_i$\Comment{{\scriptsize Sum of the independent likelihoods}}
  \EndProcedure
\end{algorithmic}
\end{algorithm}

One natural way of combining the likelihoods of the independent GPs to form the objective function of the outer loop is to take the product, which would be equivalent to taking the joint probability of the likelihoods of the independent GPs (since the likelihood is a probability density function). However, we observed that taking the sum or the harmonic mean of the likelihoods instead yielded more robust results. This comes from the fact that the product can be dominated by a few terms only and thus if some parameters explain one output dimension perfectly and all the others not as well it would still be chosen. In addition, in practice we observed that taking the sum of the likelihoods proved to be numerically more stable than the harmonic mean.

Our model learning approach, which we call \emph{GP-MI} (Gaussian Process Model Identification), that combines non-parametric model learning and parametric model identification is related to the approach in~\cite{nguyen2010using}, but there are some key differences between them. Firstly, the model learning procedure in~\cite{nguyen2010using} depends on the manipulator equation and cannot easily be used with robots that do not directly comply to the equation (one example would be the hexapod robot in our experiments or a soft robot with complex dynamics), whereas GP-MI imposes no structure on the prior model, other than providing some tunable parameters (continuous or discrete). Furthermore, the approach in~\cite{nguyen2010using} is tied to inverse dynamics models and cannot be used with forward models in the general case (necessary for long-term forward predictions); on the contrary, GP-MI can be used with inverse or forward dynamics models and in general with any black-box tunable prior model.

\subsection{Policy Search with the Black-DROPS algorithm}


We use the Black-DROPS~\cite{chatzilygeroudis2017black} algorithm for policy search because it allows us to use the type of priors discussed in Section~\ref{sec:gp_mi} and to leverage specific policy parameterizations that are suitable for different cases (\emph{e.g.}, we use a neural network policy for the pendubot task and an open-loop periodic policy for the hexapod). We assume \emph{no prior information on the policy parameters} and we begin by optimizing the policy on the prior model. Moreover, we took advantage of multi-core architectures to speed-up our experiments. Contrary to Black-DROPS, PILCO~\cite{deisenroth_gaussian_2015} cannot take advantage of multiple cores\footnote{For reference, each run of PILCO with priors (26 episodes + model learning) in the pendubot task took around 70 hours on a modern computer with 16 cores, whereas each run of Black-DROPS with priors and Black-DROPS with GP-MI took around 15 hours and 24 hours respectively.} and the need for deriving all the gradients for a different policy/reward makes it difficult (or even impossible) to try new ideas/policies.

To take the uncertainties of the model into account, the  core idea of Black-DROPS is to avoid to compute the expected reward of policy parameters, which is what most approaches do and is usually either computationally expensive~\cite{kupcsik_model-based_2014} or requires some approximation to be made~\cite{deisenroth_gaussian_2015}. Instead it treats each Monte-Carlo rollout as a noisy measurement of a function $G(\boldsymbol{\theta})$ that is the actual function $J(\boldsymbol{\theta})$ perturbed by a noise $N(\boldsymbol{\theta})$ and tries to maximize its expectation:
\begin{align}
  \mathbb{E}\Big[G(\boldsymbol{\theta})\Big]
    &= \mathbb{E}\Big[J(\boldsymbol{\theta})  + N(\boldsymbol{\theta})\Big]\nonumber
     = \mathbb{E}\Big[J(\boldsymbol{\theta})\Big] + \mathbb{E}\Big[N(\boldsymbol{\theta})\Big]\\
 & = J(\boldsymbol{\theta}) + \mathbb{E}\Big[N(\boldsymbol{\theta})\Big] (\mathrm{since~} \mathbb{E}\Big[\mathbb{E}[x]\Big] = \mathbb{E}[x])
\end{align}
We assume that $\mathbb{E}[N(\boldsymbol{\theta})] = 0$ for all $\boldsymbol{\theta}\in\mathbb{R}^{\Theta}$ and therefore maximizing $\mathbb{E}[G(\boldsymbol{\theta})]$ is
 equivalent to maximizing $J(\boldsymbol{\theta})$ (see Eq.~\eqref{eq:reward_j}).
The second main idea of Black-DROPS, is to use a population-based black-box optimizer that (1) can optimize noisy functions and (2) can take advantage of multi-core computers. Here we use BIPOP-CMAES~\cite{hansen2001completely,chatzilygeroudis2017black}. 

PI-REM~\cite{saverianodata} is close to our approach as it leverages priors to learn the residual model and then performs policy search on the model. However, PI-REM assumes that the prior information is fixed and cannot be tuned, whereas our approach has the additional flexibility of being able to change the behavior of the prior. In addition, PI-REM utilizes the policy search procedure of PILCO that can be limiting in many cases as already discussed. Nevertheless, as Black-DROPS and PILCO have been shown to perform similarly when PILCO's limitations are not present~\cite{chatzilygeroudis2017black}, we include in our experiments a variant of our approach that resembles PI-REM (Black-DROPS with priors).


\section{Experimental Results}

\subsection{Pendubot swing-up task}
\label{sec:pendubot_setup}

\begin{wrapfigure}{R}{0.45\linewidth}
  \centering
  \includegraphics[width=\linewidth]{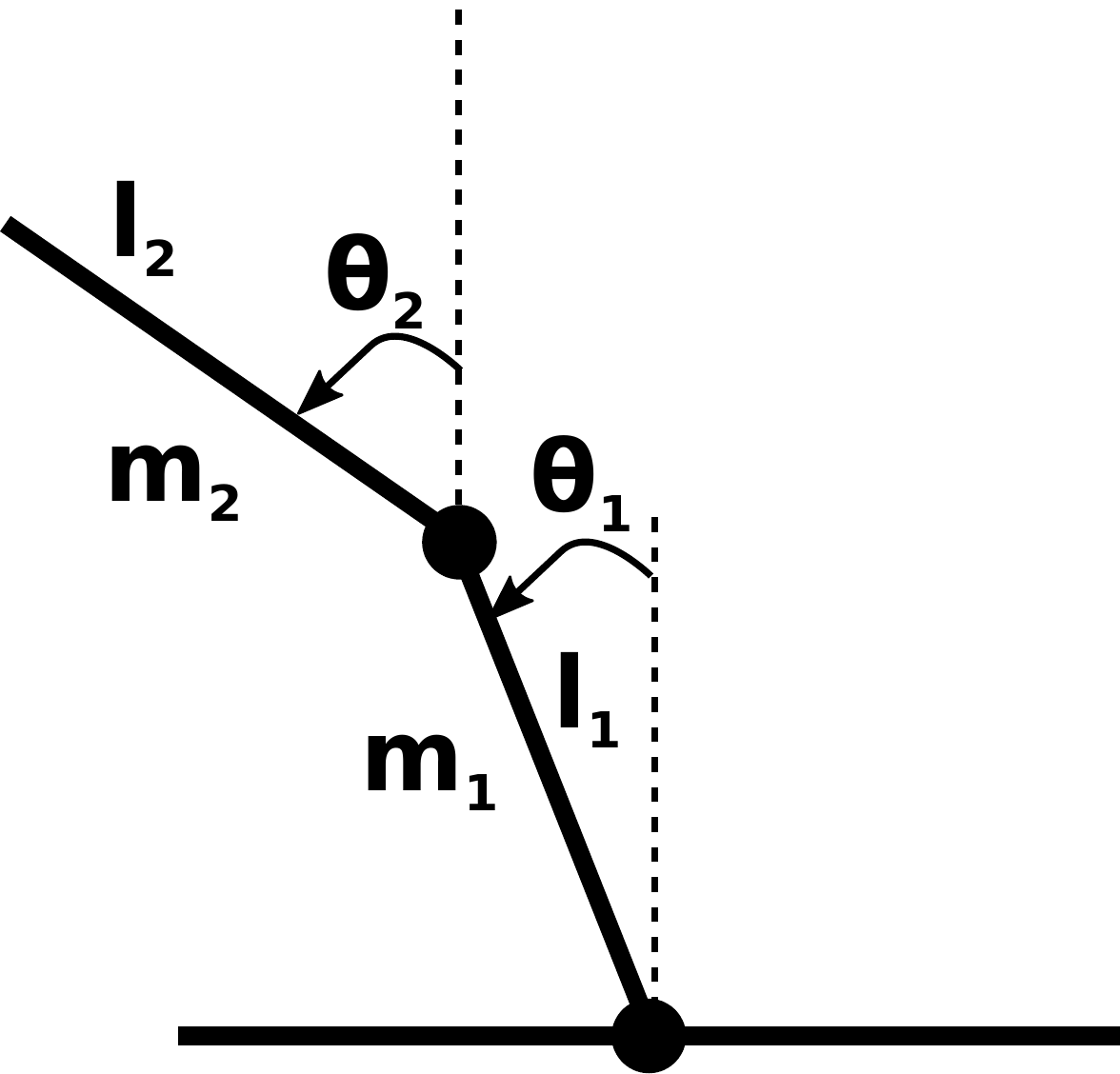}
  \vspace{-1em}
  \caption{The pendubot system}
  \label{fig:pendubot}
  \vspace{-1em}
\end{wrapfigure}

We first evaluate our approach in simulation with the pendubot swing-up task. The pendubot is a two-link under-actuated robotic arm (with lenghts $l_1$, $l_2$ and masses $m_1$, $m_2$) and was introduced by~\cite{spong1995pendubot} (Fig.~\ref{fig:pendubot}). The inner joint (attached to the ground) exerts a torque $|u|\leq 3.5$, but the outer joint cannot (both of the joints are subject to some friction with coefficients $b_1$, $b_2$). The system has four continuous state variables: two joint angles and two joint angular velocities. The angles of the joints, $\theta_1$ and $\theta_2$, are measured anti-clockwise from the upright position. The pendubot starts hanging down and the goal is to find a policy such that the pendubot swings up and then balances in the upright position. Each episode lasts $2.5$\,s and the control rate is $20$\,Hz. We use a distance based reward function as in~\cite{chatzilygeroudis2017black}.
\begin{figure*}[!ht]
  \centering
  \includegraphics[width=\textwidth]{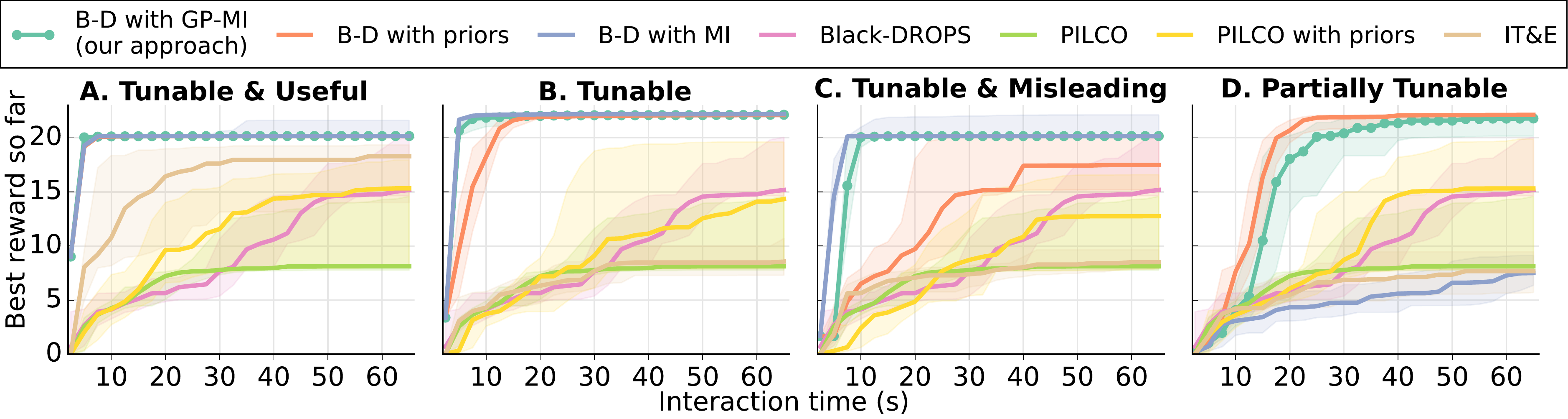}
  \vspace{-2em}
  \caption{Results for the pendubot task (30 replicates of each scenario). The lines are median values and the shaded regions the $25^{th}$ and $75^{th}$ percentiles. See Table~\ref{tab:priors} for the description of the priors. Black-DROPS with GP-MI always solves the task and achieves high rewards at least as fast as all the other approaches in all the cases that we considered. Black-DROPS with MI achieves good rewards whenever the parameters it can tune are the ones that are wrong (\textbf{A,B,C}) and bad rewards otherwise (\textbf{D}). Black-DROPS with priors performs very well whenever the prior model is not too far away from the real one (\textbf{A,B}) and not so well whenever the prior is misleading (\textbf{C}). Black-DROPS with priors and MI have very similar performance in \textbf{A} and as such are not easily distinguishable. IT\&E and PILCO with priors are not able to reliably solve the task across different prior models.}
  \label{fig:pendubot_results}
  \vspace{-1.5em}
\end{figure*}

We chose this task because it is a fairly difficult problem and forces slower convergence on model-based techniques without priors, 
but not too hard (\emph{i.e.}, it can be solved without priors in reasonable interaction time); a fact that allowed us to make a rather extensive evaluation with meaningful comparisons (4 different prior models, 7 different algorithms, 30 replicates of each combination).
We assume that we have 4 priors available; we tried to capture easy and difficult cases and cases where all the wrong parameters can be tuned or not (see Table~\ref{tab:priors}): \textbf{Tunable \& Useful:} a fully tunable prior that is very close to the actual one; \textbf{Tunable:} a fully tunable prior that is not very close to the actual; \textbf{Tunable \& Misleading:} a prior that can be fully tuned, but is very far from the actual; \textbf{Partially tunable:} a prior that cannot be fully tuned, but not very far from the actual.

We compare 7 algorithms: \textbf{1.} Black-DROPS~\cite{chatzilygeroudis2017black}; \textbf{2.} Black-DROPS with priors, which is close to PI-REM~\cite{saverianodata} and GP-ILQG~\cite{lee2017gp}\footnote{The algorithm in this specific form is first formulated in this paper (\emph{i.e.}, the Black-DROPS policy search procedure with a prior model), but, as discussed above, it is close in spirit with GP-ILQG~\cite{lee2017gp} and PI-REM~\cite{saverianodata}. Therefore, we assume that the performance of Black-DROPS with priors is representative of what could be achieved with PI-REM and GP-ILQG, although Black-DROPS with priors should be more effective because it performs a more global search~\cite{chatzilygeroudis2017black}.}; \textbf{3.} Black-DROPS with GP-MI (\emph{our approach}); \textbf{4.} Black-DROPS with MI (Black-DROPS where model learning is replaced by model identification --- via mean squared error); \textbf{5.}  PILCO~\cite{deisenroth_gaussian_2015}; \textbf{6.} PILCO with priors~\cite{cutler_efficient_2015}; \textbf{7.} IT\&E~\cite{cully2015robots}.
\begin{table}[!tb]
  \centering
  \resizebox{\columnwidth}{!}{%
	\begin{tabular}{c||c|cccc}
    \toprule
    \makecell{Variable} & \textit{Actual} & \makecell{Tunable \&\\Useful Prior} & \makecell{Tunable\\Prior} & \makecell{Tunable \&\\Misleading Prior} & \makecell{Partially\\Tunable Prior} \\
    \hline
    \makecell{$m_1$} & \textit{0.5} & \makecell{\textbf{0.65}\\\scriptsize(30\% incr.)} & 0.5 & 0.5 & \makecell{\textbf{0.65}\\\scriptsize(30\% incr.)} \\
    \hline
    \makecell{$m_2$} & \textit{0.5} & 0.5 & \makecell{\textbf{0.75}\\\scriptsize(50\% incr.)} & 0.5 & \makecell{\textbf{0.35}\\\scriptsize(30\% decr.)} \\
    \hline
    \makecell{$l_1$} & \textit{0.5} & 0.5 & 0.5 & 0.5 & 0.5 \\
    \hline
    \makecell{$l_2$} & \textit{0.5} & \makecell{\textbf{0.4}\\\scriptsize(20\% decr.)} & 0.5 & \makecell{\textbf{0.25}\\\scriptsize(50\% decr.)} & 0.5 \\
    \hline
    \makecell{$b_1$\\non-tunable} & \textit{0.1} & 0.1 & 0.1 & 0.1 & \makecell{\textbf{0.}\\\scriptsize(100\% decr.)} \\
    \hline
    \makecell{$b_2$\\non-tunable} & \textit{0.1} & 0.1 & 0.1 & 0.1 & \makecell{\textbf{0.}\\\scriptsize(100\% decr.)} \\
		\bottomrule
  \end{tabular}
  }
	\caption{Actual system and priors for the pendubot task.}
  \label{tab:priors}
  \vspace{-3em}
\end{table}

For Black-DROPS with GP-MI and the MI variant, we additionally assume that the parameters $m_1$, $m_2$, $l_1$ and $l_2$ can be tuned, but the parameters $b_1$ and $b_2$ are fixed and cannot be changed. Since the adaptation part of IT\&E is a deterministic algorithm (given the same prior) and our system has no uncertainty, for each prior we generated 30 archives with different random seeds and then ran the adaptation part of IT\&E once for each archive. We used 3 equally spread in time end-effector positions as the behavior descriptor for the archive generation with MAP-Elites. For all the Black-DROPS variants and for IT\&E we used a neural network policy with one hidden layer (10 hidden neurons) and the hyperbolic tangent as the activation function.

Similarly to IT\&E, since PILCO with priors is a deterministic algorithm given the same prior, for each prior we ran PILCO 30 times with different random seeds on the prior model (for 40 episodes in order for PILCO to converge to a good policy and model) and then ran PILCO with priors on the actual system once for each different model. We used priors both in the policy and the dynamics model when learning in the actual system (as advised in~\cite{cutler_efficient_2015}). We also used a GP policy with 200 pseudo-observations~\cite{deisenroth_gaussian_2015}\footnote{These are the parameters that come with the original code of PILCO. We used the code from: https://bitbucket.org/markjcutler/gaussian-process.}.

Black-DROPS with GP-MI always solves the task and achieves high rewards at least as fast as all the other approaches in the cases that we considered (Fig.~\ref{fig:pendubot_results}). Black-DROPS with MI performs very well when the parameters it can tune are the ones that are wrong (Fig.~\ref{fig:pendubot_results}\textbf{A,B,C}), and badly otherwise (Fig.~\ref{fig:pendubot_results}\textbf{D} --- \emph{i.e.}, no parameters of the prior model can explain the data). Black-DROPS with priors performs very well whenever the prior model is not far away from the real one (Fig.~\ref{fig:pendubot_results}\textbf{A,B}) and not so well whenever the prior is misleading (Fig.~\ref{fig:pendubot_results}\textbf{C}). Both Black-DROPS and PILCO cannot solve the task in less than $65$\,$s$ of interaction time, but Black-DROPS shows a faster learning curve (Fig.~\ref{fig:pendubot_results}).

Interestingly, PILCO with priors is not able to always achieve better results than Black-DROPS and is always worse than Black-DROPS with priors. This can be explained by the fact that PILCO without priors learns slower than Black-DROPS and is a more local search algorithm and as such needs more interaction time to achieve good results. On the contrary, Black-DROPS uses a modified version of CMA-ES that can more easily escape local optima~\cite{chatzilygeroudis2017black}. Moreover, the initial prior model for PILCO with priors is an approximated model, whereas Black-DROPS with priors uses the actual prior model to begin with. Lastly, the GP policy, that PILCO is mainly used with\footnote{So far, PILCO can only be used with linear or GP policy types~\cite{deisenroth_gaussian_2015}.}, creates really high dimensional policy spaces compared to the simple neural network policy that Black-DROPS is using (\emph{i.e.}, 1400 vs 81 parameters) and as such causes the policy search to converge slower.

IT\&E is not able to reliably solve the task and achieve high rewards. This is because IT\&E assumes that (a) the system is redundant enough so that the task can be solved in many different ways and (b) there is a policy/controller in the pre-computed archive that can solve the task (\emph{i.e.}, IT\&E cannot search outside of this archive)~\cite{cully2015robots}. Obviously, these assumptions are violated in the pendubot scenario: (a) the system is underactuated and thus does not have the required redundancy, and (b) the system is inherently unstable and as such precise policy parameters are needed (it is highly unlikely that one of them exists in the pre-computed archive).

\subsection{Physical hexapod locomotion}
\label{sec:hexa_setup}

We also evaluate our approach on the hexapod locomotion task as introduced in the IT\&E paper~\cite{cully2015robots} with a physical robot (Fig.~\ref{fig:hexapod}\textbf{A}). This scenario is where IT\&E excels and achieves remarkable recovery capabilities~\cite{cully2015robots}. We assume that a simulator of the intact robot is available (Fig.~\ref{fig:hexapod}\textbf{B})\footnote{We use the DART simulator~\cite{lee2018dart}.}; for GP-MI we also assume that we can alter this simulator by removing 1 leg of the hexapod (\emph{i.e.}, there are 7 discrete different parameterizations). This simulator is not accurate as we assume perfect velocity actuators and infinite torque. Each leg has 3 DOF leading to a total of 18 DOF. The state of the robot consists of 18 joint angles, 18 joint velocities, a 6D Center Of Mass (COM) pose (position and orientation) and 6D COM velocities. 
The policy is an open-loop controller with 36 parameters that outputs 18D joint angles every $0.1$\,s and is similar to the one used in~\cite{cully2015robots}. Each episode lasts $4$\,s and the robot is tracked with a motion capture system.

The task is to find a policy to walk forward as fast as possible. Due to the complexity of the problem\footnote{PILCO and Black-DROPS could not find any solution in preliminary simulation experiments even after several minutes of interaction time and Black-DROPS with priors was worse than Black-DROPS with GP-MI.}, we only compare 2 algorithms (IT\&E and our approach) on 2 different conditions: (a) crossing the reality-gap problem; in this case our approach cannot mostly rely on the identification part and the importance of the GP modeling will be highlighted, and (b) one rear leg is removed; the back leg removals are especially difficult 
as most effective gaits of the intact robot rely on them.

\begin{figure}[!t]
  \centering
  \includegraphics[width=0.85\linewidth]{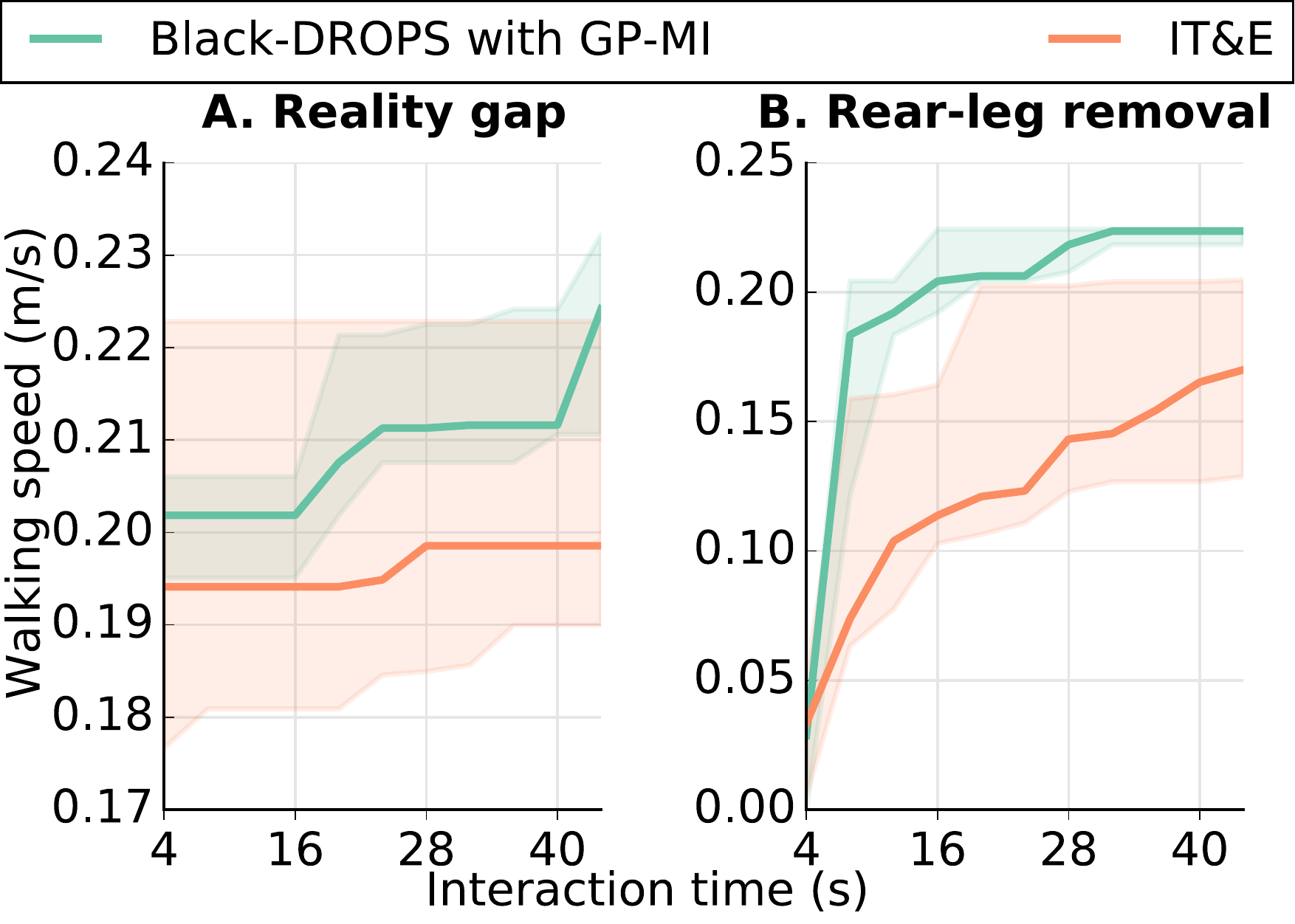}
  \caption{Results for the physical hexapod locomotion task (5 replicates of each scenario). The lines are median values and the shaded regions the $25^{th}$ and $75^{th}$ percentiles. \textbf{A.} Improving a policy for the intact robot (crossing the reality gap): Black-DROPS with GP-MI finds a highly-effective policy (about $0.22 m/s$) in less than $30$ seconds of interaction time, whereas IT\&E is not able to substantially improve the initial policy. \textbf{B.} Rear-leg removal damage case: Black-DROPS with GP-MI allows the damaged robot to walk effectively after only $16$ to $30$ seconds of interaction time and finds higher-performing policies than IT\&E ($0.21m/s$ vs $0.15m/s$ in the $8^{\text{th}}$ episode).}
  \label{fig:hexapod_results}
  \vspace{-1.7em}
\end{figure}

The results show that Black-DROPS with GP-MI is able to learn highly effective walking policies on the physical hexapod robot (Fig.~\ref{fig:hexapod_results}). In particular, using the dynamics simulator as prior information Black-DROPS with GP-MI is able to achieve better (and with less variance) walking speeds than IT\&E~\cite{cully2015robots} on the intact physical hexapod (Fig.~\ref{fig:hexapod_results}\textbf{A}). Moreover, in the rear-leg removal damage case Black-DROPS with GP-MI allows the damaged robot to walk effectively after only $16$ to $30$ seconds of interaction time and finds higher-performing policies than IT\&E ($0.21m/s$ vs $0.15m/s$ in the $8^{\text{th}}$ episode) (Fig.~\ref{fig:hexapod_results}\textbf{B}).

Overall, Black-DROPS with GP-MI was able to successfully learn working policies even though the dimensionality of the state and the action space of the hexapod robot is 48D and 18D respectively. In addition, in the rear leg damage case, Black-DROPS always tried safer policies than IT\&E that too often executed policies that would cause the robot to fall over. A video of our algorithm running on the damaged hexapod is available at the supplementary video (also at {\scriptsize\url{https://youtu.be/HFkZkhGGzTo}}).


\section{Conclusion and Discussion}
\label{sec:conclusion}

Black-DROPS with GP-MI is one of the first model-based policy search algorithms that can efficiently learn with high-dimensional physical robots. It was able to learn walking policies for a physical hexapod (48D state and 18D action space) in less than 1 minute of interaction time, \emph{without any prior on the policy parameters} (that is, it learns a policy from scratch). The black-box nature of our approach along with the extra flexibility of tuning the black-box prior model opens a new direction of experimentation as changing priors, robots or tasks requires minimum effort.

The way we compute the long-term predictions (\emph{i.e.}, by chaining model predictions) requires that predicted states (the output of the GPs) are fed back to the prior simulator. This can cause the simulator to crash because there is no guarantee that the predicted state, that possibly makes sense in the real world, will make sense in the prior model; especially when the two models (prior and real) differ a lot and when there are obstacles and collisions involved. This also holds for most other prior-based methods~\cite{lee2017gp,saverianodata,cutler_efficient_2015}, but it is not easily seen in simple systems. On the contrary, we observed this phenomenon a few times in our hexapod experiments. Using the prior simulator just as a reference and not mixing prior and real data is a direction of future work.

Finally, Black-DROPS with GP-MI brings closer trial-and-error and diagnosis-based approaches for robot damage recovery. It successfully combines (a) diagnosis~\cite{isermann2006fault} (\emph{i.e.}, identifying the likeliest robot model from data), (b) prior knowledge of possible damages/different conditions that a robot may face and (c) trial-and-error learning.





\vspace{-0.5em}
\section*{Appendix}
Code for replicating the experiments: {\scriptsize\url{https://github.com/resibots/blackdrops}}.

\vspace{-0.6em}
\section*{Acknowledgments}
The authors would like to thank Dorian Goepp, Rituraj Kaushik, Jonathan Spitz, and Vassilis Vassiliades for their feedback.


\vspace{-1em}
\bibliographystyle{IEEEtran}
\bibliography{IEEEabrv,mybib}

\end{document}